\documentclass[10pt,twocolumn,letterpaper]{article}

\usepackage[pagenumbers]{cvpr} 

\usepackage[dvipsnames]{xcolor}

\definecolor{cvprblue}{rgb}{0.21,0.49,0.74}
\usepackage[pagebackref,breaklinks,colorlinks,citecolor=cvprblue]{hyperref}

\def\paperID{6494} 
\def\confName{CVPR}
\def\confYear{2025}

\title{Edit as You See: Image-guided Video Editing via Masked Motion Modeling}

\author{Zhi-Lin Huang$^{1}$\thanks{Equal Contribution}\quad Yixuan Liu$^{1}$\footnotemark[1]\space\space\thanks{Corresponding Author}\quad Chujun Qin$^{2}$\quad Zhongdao Wang$^{3}$\quad \\ Dong Zhou$^{1}$\quad Dong Li$^{1}$ \quad Emad Barsoum$^{1}$\\
$^1$ AMD $^2$ Peking University $^3$ Tsinghua University\\
{\tt\small \{zhihuang,yixuanl,dong.zhou,d.li,ebarsoum\}@amd.com}\\ 
{\tt\small chujun.qin@pku.org.cn, \tt\small wcd17@tsinghua.edu.cn}
}

\usepackage{amsmath,amsfonts,bm}

\def\eqref#1{equation~\ref{#1}}

\def\1{\bm{1}}

\DeclareMathAlphabet{\mathsfit}{\encodingdefault}{\sfdefault}{m}{sl}
\SetMathAlphabet{\mathsfit}{bold}{\encodingdefault}{\sfdefault}{bx}{n}

\def\gD{{\mathcal{D}}}
\def\gE{{\mathcal{E}}}
\def\gF{{\mathcal{F}}}

\def\gI{{\mathcal{I}}}

\def\gL{{\mathcal{L}}}
\def\gM{{\mathcal{M}}}
\def\gN{{\mathcal{N}}}
\def\gO{{\mathcal{O}}}

\def\gV{{\mathcal{V}}}

\def\gZ{{\mathcal{Z}}}

\usepackage{pifont}

\usepackage{stmaryrd}
\usepackage{amsmath}
\usepackage{amssymb}
\usepackage{mathtools}
\usepackage{amsthm}
\usepackage{dsfont}
\usepackage{diagbox}
\usepackage{adjustbox}
\usepackage{multirow}
\usepackage{makecell}
\usepackage{graphicx}
\usepackage{caption}
\usepackage{booktabs}
\usepackage{cellspace}
\usepackage{pifont}
\usepackage{amssymb}
\usepackage{tabularx}

\theoremstyle{plain}

\theoremstyle{definition}

\theoremstyle{remark}

\usepackage{xspace}
\usepackage[capitalize,noabbrev]{cleveref}
\crefname{section}{Sec.}{Secs.}
\Crefname{section}{Section}{Sections}
\Crefname{table}{Table}{Tables}
\crefname{table}{Tab.}{Tabs.}
\usepackage{xspace}
\usepackage{yhmath}
\usepackage{pifont}
\newcommand{\method}{\textsc{IVEDiff}\xspace}
\newcommand{\module}{\textsc{MotRefNet}\xspace}
\newcommand{\mmm}{\textsc{MMM}\xspace}

\usepackage{algorithm}
\usepackage{listings}

\definecolor{codegreen}{rgb}{0,0.6,0}
\definecolor{codegray}{rgb}{0.5,0.5,0.5}
\lstdefinestyle{mystyle}{
    commentstyle=\color{codegreen},
    keywordstyle=\color{magenta},
    numberstyle=\tiny\color{codegray},
    stringstyle=\color{codepurple},
    basicstyle=\ttfamily\footnotesize,
    breaklines=true,                 
    keepspaces=true,                 
    numbers=left,                    
    showstringspaces=false,
}
\begin{document}

\maketitle

\begin{abstract}
Recent advancements in diffusion models have significantly facilitated text-guided video editing. However, there is a relative scarcity of research on image-guided video editing, a method that empowers users to edit videos by merely indicating a target object in the initial frame and providing an RGB image as reference, without relying on the text prompts. 
In this paper, we propose a novel \textbf{I}mage-guided \textbf{V}ideo \textbf{E}diting \textbf{Diff}usion model, termed \textbf{\method} for the image-guided video editing. 
\method is built on top of image editing models, and is equipped with learnable motion modules to maintain the temporal consistency of edited video.
Inspired by self-supervised learning concepts, we introduce a masked motion modeling fine-tuning strategy that empowers the motion module's capabilities for capturing inter-frame motion dynamics, while preserving the capabilities for intra-frame semantic correlations modeling of the base image editing model.
Moreover, an optical-flow-guided motion reference network is proposed to ensure the accurate propagation of information between edited video frames, alleviating the misleading effects of invalid information.
We also construct a benchmark to facilitate further research. The comprehensive experiments demonstrate that our method is able to generate temporally smooth edited videos while robustly dealing with various editing objects with high quality. 

\end{abstract}
\section{Introduction}

In recent years, video, as a medium for information transmission, has not only changed the way information is disseminated but also profoundly influenced human communication, learning, and entertainment, becoming an essential part of modern society. With the rapid growth of video content on the Internet, the demand of users for video editing has also increased dramatically. 

Recently, diffusion models~\cite{ho2020denoising,song2021scorebased,ho2022video,rombach2022high,dhariwal2021diffusion,compvis}, a powerful generative model, have shown promising progress in the fields of image and video generation, and have demonstrated significant potential in video editing tasks~\cite{shi2024motion,ceylan2023pix2video,liu2024video,chai2023stablevideo,geyer2023tokenflow,qi2023fatezero,couairon2023videdit,wang2023zero,jeong2024vmc}.By automatically editing the given video, the costs of manual editing can be greatly reduced.
A series of works~\cite{yan2023motion,yan2023magicprop,zhao2025motiondirector} firstly use pre-trained text-to-image models to achieve the first video frame editing, and then using image animation models to predict subsequent frames, thereby generating new video content that aligns with the given editing text prompt. For preserving the motion trajectories and spatial structures of the original video, attention manipulations and the motion conditions (like object trajectories, etc.) are always incorporated.
Unfortunately, these methods can only change the global style of the entire video, and are generally unable to perform local fine-grained editing without modifying unrelated regions. Meanwhile, most of the current work focuses on achieving video editing based on text descriptions. However, \textbf{\textit{a picture is worth a thousand words}}, only a few words are insufficient to describe rich content. By decomposing the video editing task into indicating contents that to be edited in the source video and providing an image as the reference of edited contents, models can generate edited videos that more closely align with the user's imagination. 
Recently, MimicBrush~\cite{chen2024zero} has been proposed as a powerful image-guided image editing (IIE) diffusion model, which has capabilities of capturing semantic correlations between two images, enabling direct editing of contents within the marked regions of the source image based on the reference one. This provides a powerful tool for subsequent research on image-guided video editing (IVE).

To implement IVE based on off-the-shelf IIE models, a straightforward approach is to apply the IIE method to edit each frame of the source video directly. However, this method does not take into account the temporal consistency of the edited frames, making it difficult to generate visually smooth results. 
Meanwhile, training a video editing diffusion model from scratch is resource-intensive.
Thus, a more reasonable and resource-saving method is to utilize image animation techniques~\cite{wang2022latent,xing2025dynamicrafter,molad2023dreamix}. 
AnimateDiff~\cite{guo2023animatediff}, a promising image animation method, first inflates an image diffusion model and then integrates a learnable motion module to capture inter-frame motion dynamics, followed by fine-tuning the motion module using the original de-noising objectives.
However, directly applying the fine-tuning strategy used in AnimateDiff to IVE is non-trivial for the following reasons:
\textbf{(1)} Compared to the image animation task, in the editing task, the model's input is incomplete, making the inter-frame regional correlations which is calculated directly from the corrupted regions misleading. This makes it more challenging for the model to establish reliable inter-frame correlations, leading to distorted structures.
\textbf{(2)} Directly fine-tuning the motion modules with the de-noising training objectives may cause the model to over-focus on the temporal consistency maintaining, while sacrificing the original capabilities for modeling intra-frame semantic correlations of base image editing model, resulting in an excessive loss of visual quality in each edited frame.

To this end, we propose \method, which, to the best of our knowledge, is the first image-guided video editing diffusion model. \method utilizes a powerful image-guided image editing diffusion model, MimicBrush~\cite{chen2024zero}, as the base model and incorporates an optical-flow-guided motion reference network, termed as \module, into each layer of the de-noising network to maintain the appearance consistency between neighboring edited frames under the guidance of optical flow predicted from the source video. By fully leveraging the inter-frame motion priors within the source video, \method mitigates the inaccurate correlation modeling caused by invalid information in the regions to be edited. 
Following AnimateDiff, we only fine-tune the \module in IVEDiff while keeping the other model weights fixed. Unlike AnimateDiff, to better assist the model in modeling inter-frame correlations and to minimize the degradation of the visual quality of single-frame editing results, we propose a masked motion modeling (\mmm) fine-tuning strategy. During the fine-tuning process, a randomly selected video clip is first temoprally downsampled at a certain stride, then the first video frame is taken as the reference image, and the remaining frames are partially masked at a certain ratio as the training video. Finally, the model is optimized through a de-noising objective~\cite{ho2020denoising}. With the \mmm fine-tuning strategy, \module is required to have a stronger capability for maintaining inter-frame appearance consistency. 
Besides, by aligning the fine-tuning and inference processes, \method is able to preserve the capabilities for intra-frame semantic correlation modeling of the original image-guided image editing diffusion model, thereby minimizing the loss of visual editing quality in each individual frame.
We also construct a benchmark to facilitate further research. The comprehensive experiments demonstrate that our method is able to generate temporally smooth edited videos while robustly dealing with various editing objects with high quality. The main contributions of our work are as follows:
\begin{itemize}
    
    \item We propose an optical-flow-guided reference network (\module) that incorporates the inter-frame optical flow priors from the source video. By leveraging \module, our model is capable of more effectively modeling inter-frame correlations, thereby achieving visually smooth results.

    \item For image-guided video editing, we propose an effective fine-tuning strategy, termed as masked motion modeling (\mmm). This strategy involves utilizing the initial frame of a video clip as a reference image and randomly masks spatial areas of subsequent frames. 
    By utilizing \mmm, we not only align the fine-tuning and inference processes to empower \method with the capability to model inter-frame consistency, but also maintain the ability for intra-frame semantic correlation modeling of the original image editing model, thereby minimizing the sacrifice of visual quality due to temporal consistency constraints.

    \item We have conducted a benchmark comprising five metrics to assess inter-frame consistency and frame image quality, complemented by a dataset of 236 video-mask-reference triplets. This benchmark is designed to thoroughly evaluate the efficiency of various methods in preserving visual coherence and quality across frames during image-guided video editing. The comprehensive experiments demonstrate that our method is able to generate temporally smooth edited videos while robustly dealing with various editing objects with high quality. 
\end{itemize}

\section{Related work}
\noindent \textbf{Diffusion-based Image Editing. }
Image editing models~\cite{kawar2023imagic,couairon2022diffedit,zhang2023sine,yang2023paint,shi2024dragdiffusion} typically edit the source image based on the given prompts. In addition to being required to maintain semantic consistency between the editing results and the given prompts, the editing models are sometimes also required to preserve specific characteristics of the source image, for example spatial layout.
Many attempts have been made to achieve this task using the pre-trained text-to-image model. 
IP-Adapter~\cite{ye2023ip} employs a decoupled cross-attention mechanism to integrate image prompts into pre-trained text-to-image diffusion models, preserving the original appearance of original image in the editing one.
EditAnything~\cite{gao2023editanything} leverages segmentation masks and ControlNet~\cite{zhang2023adding} to enable flexible and high-quality image editing while preserving original layout.
Recently, a pioneer image-guided image editing diffusion models, MimicBrush~\cite{chen2024zero}, is proposed. By modeling the semantic correlations between the editing image and the given reference one during the training process, the MimicBrush is able to edit the local regions according to the reference without relying on text prompts.
In our \method, we utilize the MimicBrush as our base image editing model.

\noindent \textbf{Image Animation.}
Currently, image diffusion models have achieved promising progress in various fields of computer vision. Meanwhile, due to the limitations of computational resources, it is not easy to train a video diffusion model from scratch. Therefore, image animation~\cite{wang2022latent,xing2025dynamicrafter,molad2023dreamix,singer2022make,kim2022diffusionclip} has received widespread attention recently. Most existing methods achieve image animation by leveraging the strong generative priors of image diffusion models and estimating the potential motion relationships of single-frame images.
AnimateDiff~\cite{guo2023animatediff} firstly inflates a pre-trained image model to process video data and then optimizing a temporal motion module on real-world videos while keeping the base image model weights fixed to learn generalized motion priors. GID~\cite{li2024generative} introduces spectral volumes, which are predicted by a frequency-coordinated latent diffusion model and used to animate future video frames through an image-based rendering module.
However, directly applying image animation techniques to build video editing diffusion models on top of image editing ones is non-trivial, as the modeling of motion dynamics between edited frames can be misled by the lack of input information, and models tend to sacrifice visual quality of individual frame to meet the requirements of inter-frame temporal consistency. To address these issues, we propose a \module and a \mmm fine-tuning strategy, which will be detailed below.

\noindent \textbf{Text-guided Video Editing.}
Text-guided video editing~\cite{kawar2023imagic,mokady2023null,nichol2021glide,brooks2023instructpix2pix,bar2022text2live,avrahami2022blended} aims to modify the specific foreground or background of the source videos according to the given text prompts while preserving some characteristics of the original video. 
Existing approaches inflating the pre-trained text-to-image models and incorporate the cross-frame temporal correlation modeling for achieving zero-shot text-guided video editing. 
Rerender-A-Video~\cite{yang2023rerender} is able to produces temporal coherent edited frames by applying heavy cross-frame constraints, but is limited to global style transfer and do not support local region editing. 
AVID~\cite{zhang2024avid} achieving fine-grained control by injecting the textual information into the de-noising UNet following the ControlNet.
Different from these text-guided video editing or inpainting methods, our \method can edit the specific fine-grained regions of the source video by automatically referring to the semantic-related contents in the given reference image, and the editing process is not relying on text descriptions.
\begin{figure*}[!t]
\centering
\includegraphics[width=0.99\linewidth]{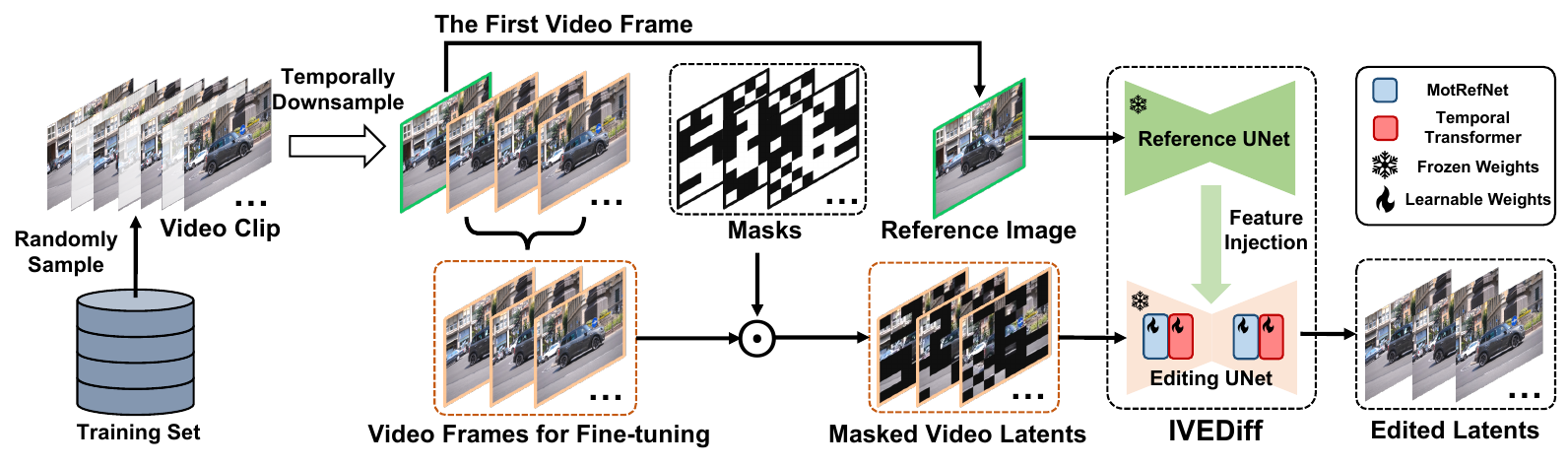}
\caption{Overall pipeline of the proposed masked motion modeling fine-tuning strategy. To align the fine-tuning process with the inference process, we use the first frame of the video clip in the training set as the reference image, and partially obscure the remaining frames. By this fine-tuning process, model gains the capabilities of capturing inter-frame temporal consistency while preserving the base image model's original ability to model intra-frame semantic correlations.
}
\label{fig:mmm}
\end{figure*}

\section{Preliminary}
We introduce the preliminary of Stable Diffusion (SD)~\cite{rombach2022high} and AnimateDiff~\cite{guo2023animatediff} for helping better understand the \module and \mmm proposed in our \method.

\noindent \textbf{Stable Diffusion.} 
Stable Diffusion (SD) is a text-to-image generation model that can automatically create high-quality, high-resolution images based on any given text.
And it is open-sourced and has a well-developed community with many high-quality variant models. 
SD operates on the principle of diffusion models which are designed to generate new data similar to what they have learned.
SD works by first compressing images into a latent space, which is significantly smaller than the pixel space, making the diffusion process much faster than traditional diffusion models that operate directly on pixel data. SD uses a variational auto-encoder (VAE) to achieve this compression, with the encoder $\gE(\cdot)$ reducing the image into a lower-dimensional representation in the latent space and the decoder $\gD(\cdot)$ reconstruct the image from the latent.
And then SD performs the diffusion process on the latent space. In the training process, an encoded image $z_0 = \gE(\gI)$ is perturbed to $z_t$ by the forward diffusion:
\begin{align}
\footnotesize
    z_t = &\sqrt{\bar{\alpha_t}}z_0 + \sqrt{1 - \bar{\alpha_t}}\epsilon \label{eq-sd-forward}
\end{align}
where $\epsilon\sim\gN(0, I)$, $t={1, ..., T}$, and $\bar{\alpha_t}$ is pre-defined and it determines the noise strength in time step $t$. The de-noising network with learnable parameters $\theta$ is optimized to reverse this diffusion process by predicting the added noises. The training objectives is:
\begin{align}
\footnotesize
    \gL = \mathbb{E}[\Vert\epsilon -& \epsilon_{\theta
    }(z_t, t, f_{prompt}(\text{text\_prompt}))] \label{eq-sd-loss}
\end{align}
where $\text{text\_prompt}$ is the given text prompt corresponding to the image $\gI$, $f_{prompt}(\cdot)$ is the encoder mapping the text prompt to a vector sequence, $\epsilon_{\theta}$ is a UNet with learnable weights $\theta$. The UNet consisting of pairs of down/up  blocks at four resolution levels, as well as the middle block. Each block consists of ResNet, spatial self-attention layers, and cross-attention layers that introduce text conditions.

\noindent \textbf{AnimateDiff.} 
Directly training a Text-to-Video (T2V) model from scratch requires a significant amount of computational resources. In contrast, AnimateDiff is a method that leverages a pre-trained Text-to-Image (T2I) model and uses minimal computational resources to achieve T2V capabilities . Specifically, rather than generating frames individually, AnimateDiff inflates the T2I model at first and then embeds a motion model into each layer of the de-noising network to construct the correlations of the same spatial location across different frame of the same video clip. This motion model is typically a temporal Transformer block which is adapted from the original Transformer one to operate information propagation along the temporal dimension. Finally, all pre-trained T2I model parameters in AnimateDiff are fixed, and the de-noising objective following the original diffusion model is used to fine-tune the motion module.

\section{Methods}
\subsection{Overall Pipeline}

\begin{figure}[!t]
\centering
\includegraphics[width=0.99\linewidth]{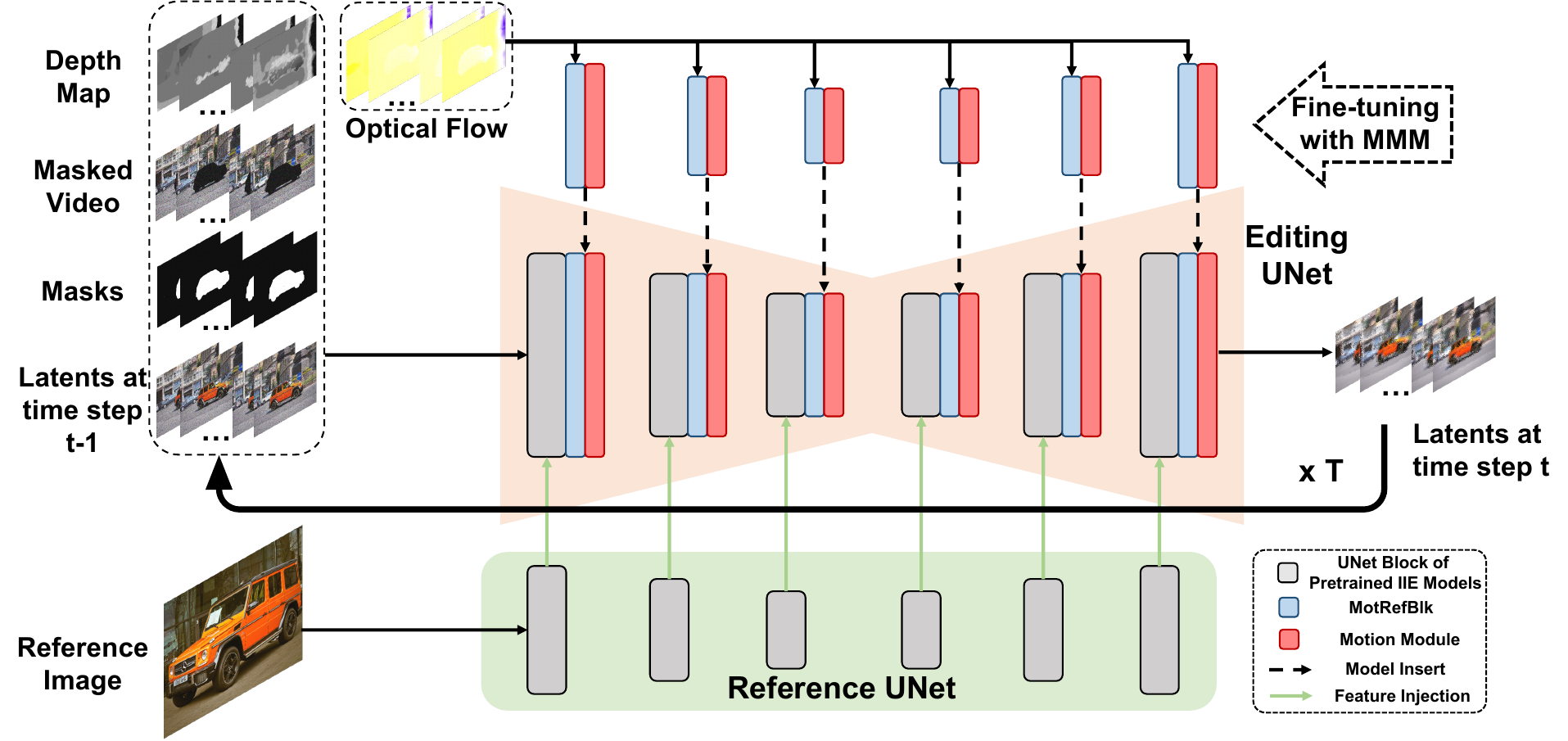}
\caption{Overall framework of the proposed \method.
}
\label{fig:arch}
\end{figure}

Compared to directly constructing a video diffusion model, we choose to leverage existing powerful image-based models to accomplish the spatial dimension generation of videos, and incorporate motion modules to model the inter-frame dynamic motion along the temporal dimension. 
As shown in~\cref{fig:arch}, the input of \method consist of a $f$-frame source video $\gV^{[0:f]}=\{\gI^0, ... \gI^{f-1}\}, \gI^i\in\mathbb{R}^{3\times U\times V}$ and a mask sequence $\gM^{[0:f]}=\{m^0, ... m^{f-1}\}, m^i\in\mathbb{R}^{U\times V}$ which indicated the target object to be edited in each frame. For the given source video, we firstly extract the depth map $\gD^{[0:f]}=\{d^0, ... d^{f-1}\}, d^i\in\mathbb{R}^{4\times U\times V}$ for each frame through the pre-trained depth estimator Depth-Anything~\cite{yang2024depth}, and extract the optical flow prior $\gO\gF^{[0:f-1]}=\{of^{0\shortrightarrow 1}, ... of^{f-2\shortrightarrow f-1}\}, of^{i\shortrightarrow i+1}\in\mathbb{R}^{2\times U\times V}$ from two neighboring frames through the pre-trained optical flow predictor GMFlow~\cite{xu2022gmflow}. And then, we frame-wisely encode the source video through the encoder of auto-encoder of pre-trained SD1.5, obtaining the encoded latents $\gZ^{[0:f]}=\{z^0, ... z^{f-1}\}, z^i\in\mathbb{R}^{4\times H\times W}$ of each frame. 
For the spatial editing, we build up our \method on the top of the MimicBrush~\cite{chen2024zero}, a powerful image-guided image editing methods. The de-noising network consists of two main components: the editing UNet and the reference UNet. Specifically, the editing UNet takes 4 parts and concatenate them along the channel dimension as input: (1) the masked latent, which is obtained by element-wisely multiplying the encoded latent and the corresponding mask; (2) the noised latent at timestep $t$; (3) the mask; (4) the depth map, where the mask and the depth map are spatially downsampled to the resolution of the noised latent.
And the reference image which is encoded to the image embedding through the CLIP, is fused into the UNet via the cross-attention mechanism.
The noised latents takes charge of the diffusion procedure from initial noises to the output latent codes step by step. 
Besides, to help the model to better capture the fine-grained information of the given reference image, the reference UNet takes the latent code of the reference image as input, where the latent code is encoded through the encoder of Auto-encoder of the SD1.5. Then the multi-level fine-grained features extracted from the reference UNet are injected into the editing UNet.
Instead of edit each frame individually, the image diffusion model inflated with learnable optical-flow-guided motion reference network (\module) and motion module learns to capture temporal changes of video frames, constituting the motion dynamics of edited target.

\subsection{Motion Reference Network}
In editing and inpainting tasks, since the regions to be edited in the input video is missing, the inter-frame correlations calculated directly from the corrupted regions are misleading. This makes it difficult for the editing model to accurately model inter-frame correlations, resulting in distorted structures. To address this issue, we propose an optical-flow-guided motion reference network, \module, which leverages the undamaged inter-frame optical flow priors from the source video which is available, to effectively guide the information propagation process across neighboring frames in the edited video.

In \module, we utilize the pre-trained GMFlow~\cite{xu2022gmflow}, which is a robust optical flow predictor, to capture the complete and confident optical flows across neighboring frames of the source video, and treat it as the information propagation prior between two neighboring frames in the edited video. 
Specifically, given two neighboring frames $\gI_{src}^{i}, \gI_{src}^{i+1}\in\mathbb{R}^{3\times U\times V}$ from the source video $\gV_{src}$, where $\gI_{src}^{i}$ denotes the previous frame and $\gI_{src}^{i+1}$ denotes the subsequent one, the uni-directional optical flow prior is firstly extracted from the pre-trained GMFlow $f_{OF}(\cdot)$:
\begin{align}
    of^{i\shortrightarrow i+1} &= f_{OF}(\gI_{src}^{i}, \gI_{src}^{i+1}) \label{eq:of}
\end{align}
where $of^{i\shortrightarrow i+1}\in\mathbb{R}^{2\times U\times V}$ denotes the optical flow indicating the pixel mapping from the previous frame $\gI_{src}^{i}$ to the next one $\gI_{src}^{i+1}$. 
Then, for the input latents $h_{edit}^{i}, h_{edit}^{i+1}\in\mathbb{R}^{C\times H\times W}$ of the corresponding frames in the editing video, a learnable neural network $f_{offset}(\cdot)$ is applied to learn the pixel-level feature correlations between two editing frames under the guidance of optical flow prior extracted from the source video:
\begin{align}
    \Omega^{i\shortrightarrow i+1} &= f_{offset}(h_{edit}^{i}, h_{edit}^{i+1}, of_{down}^{i\shortrightarrow i+1}) \label{eq:offset}
\end{align}
where $\Omega^{i\shortrightarrow i+1}\in\mathbb{R}^{2\times H\times W}$ is a 2-channel offset map, and $of_{down}^{i\shortrightarrow i+1}\in\mathbb{R}^{2\times H\times W}$ is the optical-flow map down-sampled to be aligned with the resolution of corresponding latents. After that, a warp function $f_{warp}(\cdot)$~\cite{hui2018liteflownet} is introduced and serves as a information propagation mechanism from the latent of the previous frame to the next one:
\begin{align}
\footnotesize
    h_{edit}^{i\shortrightarrow i+1} = f_{warp}(\Omega^{i\shortrightarrow i+1}, h_{edit}^{i})
\end{align}
In addition, a learnable scalar value $\alpha$ is introduced to further enhance performance and account for occlusion:
\begin{align}
\footnotesize
    \Tilde{h}_{edit}^{i+1} &= h_{edit}^{i\shortrightarrow i+1} + \alpha\cdot \delta_{edit}^{i+1}
\end{align}
where $\Tilde{h}_{edit}^{i+1}$ is the output of \module.

Finally, for a video clip of length $L$, we can obtain a sub-clip of length $L-1$ enhanced by \module, which uses the previous frame to enhance the subsequent one. Since the first frame lacks a previous frame, we directly use the original latent of the first frame as the enhanced result of first frame. By hierarchically applying \module at each layer of the de-noising network, we can implicitly achieve information propagation between frames with any interval.

\subsection{Masked Motion Modeling Fine-tuning Strategy}

Directly using the de-noising training objectives of diffusion models to fine-tune the motion module and \module not only fails to effectively exploit the potential of the motion module in modeling inter-frame semantic correlations, but also this fine-tuning process is inconsistent with the practical inference phase, severely impacting the frame-wise editing results of the original image-guided image editing model.
That is, in order to maintain temporal consistency, the model will excessively sacrifice the visual quality of the editing in each individual frame.

To address the aforementioned issue, inspired by self-supervised learning~\cite{he2022masked,tong2022videomae}, we propose a simple yet effective fine-tuning strategy, called masked motion modeling (\mmm), as shown in~\cref{fig:mmm}. This strategy aims to enhance the temporal consistency of the editing results while preserving the quality of individual frame as much as possible.
Specifically, during the fine-tuning phase, we randomly select a video from the training set and down-sample it along the temporal dimension at a certain stride to obtain a video clip of a fixed length. To maintain consistency with the model's inference phase, we choose the first frame of this video clip as the reference image and use the remaining frames for the fine-tuning. 
We feed the complete video frames into the pre-trained SD auto-encoder for frame-wisely encoding, obtaining the latent codes $Z_0$. And then the latent codes are noised using the defined forward diffusion schedule as in~\cref{eq-noise}.
Subsequently, for the latent code of each frame image in the fine-tuning process, we generate a mask to accomplish the random masking of a certain proportion of the spatial information.
For the mask generation, we create a tensor with the same resolution as the latent code, filled entirely with the value one, and divide it into an $N\times N$ grid along the spatial dimension.
Then, we element-wisely multiply the obtained masks with the latent codes, completing the random occlusion of spatial regions in the latents.
Finally, we feed the partially occluded latents into the de-noising network, simultaneously completing the noise prediction for both the occluded and un-occluded regions. And we utilize the original de-noising objective for the fine-tuning process.

Following AnimateDiff~\cite{guo2023animatediff}, we only fine-tune the motion module and \module, while keeping all other model weights in \method fixed. By randomly occluding regions in the video frames, we can effectively help our motion model and \module to model inter-frame temporal and semantic correlations, thereby enabling \method to generate visually continuous results between frames while ensuring the visual quality of single-frame editing results. The process is formulated as follows:
\begin{align}
\footnotesize
    I_{ref} = &V_{src}^{[0]},      V_{edit} = V_{src}^{[1:f+1]} \\
    z_0^{[0:f]} &= f_{enc}(V_{edit}) \\
    z_t^{[0:f]} = &\sqrt{\bar{\alpha_t}}z_0^{[0:f]} + \sqrt{1 - \bar{\alpha_t}}\epsilon^{[0:f]} \\ \label{eq-noise}
    \gL = \mathbb{E}[\Vert\epsilon -& \epsilon_{\theta
    }(z_t^{[0:f]}\odot m^{[0:f]}, t, f_{clip}(I_{ref}))]
\end{align}
where $\epsilon \sim \gN(0, I)$ is a standard Gaussian noise, $f$ is the number of frames in the editing video, $f_{enc}(\cdot)$ is the encoder of the auto-encoder of a pre-trained SD, $\theta$ is the parameter of the de-noising UNet of \method.

\section{Experiments}
\subsection{Implement Details}
\noindent \textbf{Training Details.}
We implement \method upon MimicBrush~\cite{chen2024zero} and fine-tune the \module and motion module using the Pexels dataset~\cite{opensoraplan}. 
In the masked motion modeling fine-tuning process, we set the video clip stride to 4, video frame resolution to the 512$\times$512. We choose the grid mask for masking, and set the mask grid size to 8 (the resolution of input latent is 64$\times$64, the mask ratio to 0.5.
We use the AdamW as the optimizer, set the learning rate to $10^{-5}$. The type of learning rate scheduler is constant. 
And we follow DDPM~\cite{ho2020denoising}, and use 1000 steps for the noising schedule.
For the \module, we initialize all layers with zero. 
For each layer of the de-noising network, we embed an additional motion module following the embedded \module, and we initialize all layers of the motion module with AnimateDiffV3~\cite{guo2023animatediff}.
For the other layers of the de-noising network, we initialize with the pre-trained weights from MimicBrush~\cite{chen2024zero}, and ensure that only the weights of \module and the motion module are optimized during the fine-tuning process. 

\noindent \textbf{Inference Details.}
In the inference stage, we follow DDIM~\cite{song2020denoising}, use a 50 sampling steps and the classifier-free guidance scale is 7.5. The mask for each video frame can be obtained by any automatic mask generation tools, for example, SAM2~\cite{kirillov2023segment} and Xmem~\cite{cheng2022xmem}, or provided by the user according to their preferences and intentions. 
To evaluation the performance of our \method in image-guided video editing, we conduct our experiments on the IVE-Benchmark we proposed. The details of IVE-Benchmark will be introduced below.

\noindent\textbf{IVE-Benchmark.}
Due to the lack of relevant prior work, we construct our own benchmark to systematically evaluate the model's performance in editing specified areas of a video by referring to a given reference image.
We divide the application into two sub-applications: object modification and texture transfer. Additionally, we provide two levels of granularity for masks: fine-grained object masks and coarse-grained rectangular masks. The fine-grained masks are obtained by segmenting the specified objects in each frame using SAM2, while the coarse-grained masks are created by building the smallest rectangle that can fully cover the fine-grained masks. We use the fine-grained masks for both object modification and texture transfer applications, and the coarse-grained masks are used for the object modification application.
Ultimately, we select 40 source videos from the Davis90 dataset and extracted masks for 101 objects and backgrounds. And the objects covering four categories: animals, clothing, vehicle, and daily items. Additionally, for both object modification and texture transfer applications, we respectively provided reference images of the different and the same types of objects for the videos to be edited. In the end, we obtained a total of 236 \textbf{\textit{video-object-reference}} triplets.
Additionally, we assess the performance of different methods from 3 perspectives, including the temporal smoothness (Warp Error~\cite{lai2018learning}, Temporal Consistency~\cite{zhang2024avid}), the visual quality of edited frames (FID~\cite{Seitzer2020FID}) and the semantic alignment between edited frames and the reference image (CLIP Score~\cite{taited2023CLIPScore}).

\begin{figure*}[!t]
\centering
\includegraphics[width=0.99\linewidth]{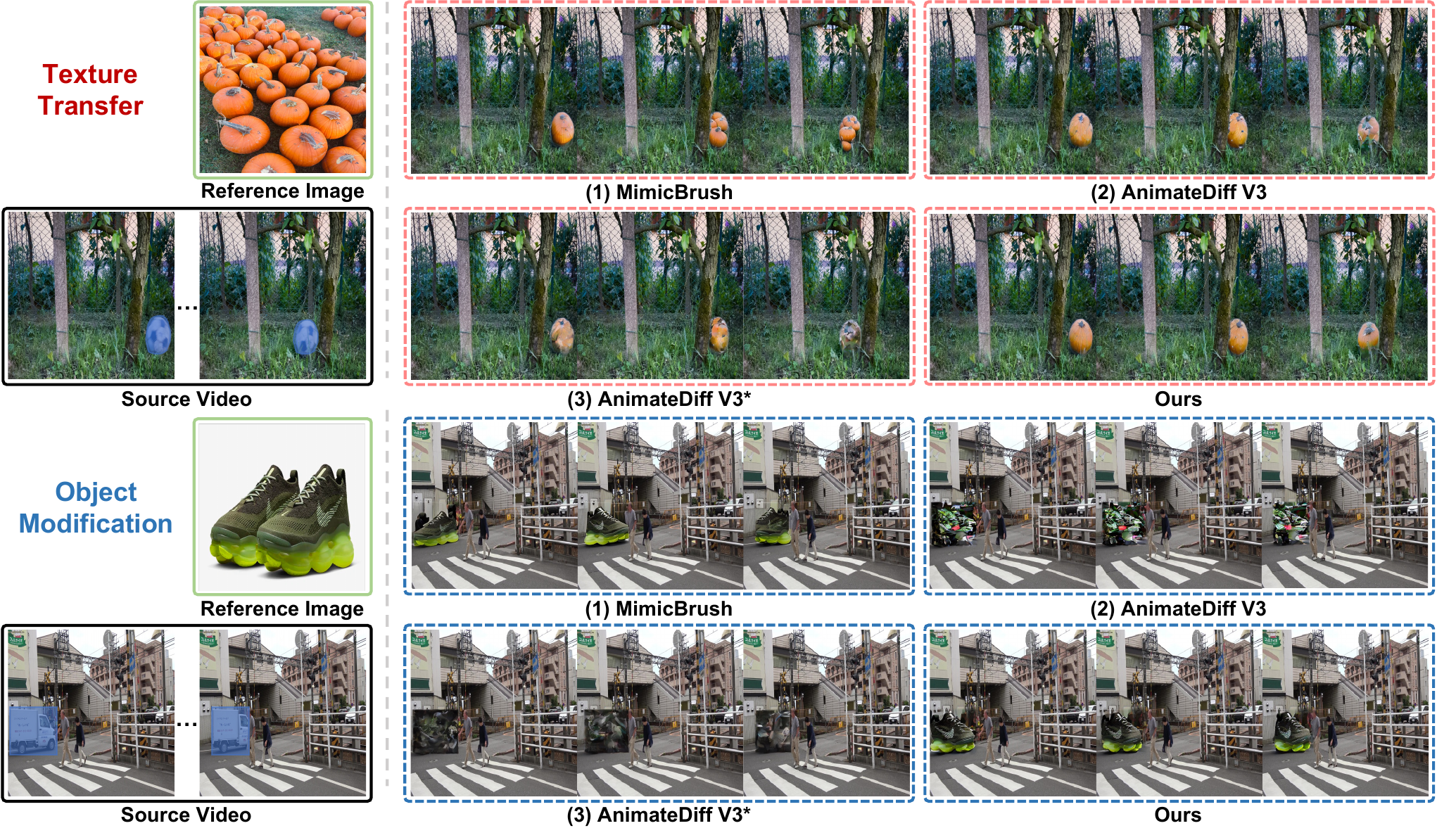}
\caption{The qualitative comparisons between our methods with baselines. For the texture transfer, we use the fine-grained masks which are generated by SAM2 to indicate the editing target, and use the highly semantic-related reference image; For the object modification, we use the rectangle masks which coarsely cover the editing target and the semantically unrelated reference image.
}
\label{fig:qualicomp}
\end{figure*}

\subsection{Quantitative Comparison.}

Due to the deficiencies of previous IVE methods, we have selected two models as baselines to compare with our \method: (1) MimicBrush~\cite{chen2024zero}, which edits each frame of the source video frame-wisely based on the given reference image; (2) AnimateDiffV3~\cite{guo2023animatediff}, whose implementation process first involves expanding MimicBrush and then using the pretrained AnimateDiffV3 motion module to model inter-frame consistency; (3) AnimateDiffV3$^\star$, the motion module of AnimateDiffV3 is fine-tuned on our training dataset.
According to the IVE-Benchmark, we evaluate the performance of the IVE models in two applications: texture transfer and object modification. The evaluation results are shown in~\cref{tab:quanticomp}. 
We can observe that our \method significantly outperforms other baselines in maintaining temporal consistency. Besides, in terms of video frame visual quality and the semantic alignment between the edited results and the reference image, the model may make some sacrifices due to the additional inter-frame consistency constraints. For the texture transfer, our \method incurs some loss compared to frame-wise editing, but achieves a better trade-off between image quality and inter-frame consistency compared to other baselines. Meanwhile, for object modification, our \method actually improves the visual quality of the editing results compared to the baseline, which we speculate is due to the consistency constraints stabilizing large-region editing results and avoiding the generation of extremely poor editing results by the model during the frame-wise editing.

\begin{table*}
\centering
\begin{adjustbox}{width=0.95\textwidth}
    \renewcommand{\arraystretch}{1.2}
\begin{tabular}{lcccclcccc}
\toprule
\multirow{2}{*}{Methods} &
\multicolumn{4}{c}{Texture Transfer} & & \multicolumn{4}{c}{Object Modification}
\\ \cmidrule{2-5} \cmidrule{7-10}
& Warp Error ($10^{-1}$) $\downarrow$ & Temporal Consistency $\uparrow$ & FID $\downarrow$ & CLIP Score $\uparrow$ & & Warp Error ($10^{-1}$) $\downarrow$ & Temporal Consistency $\uparrow$ & FID $\downarrow$ & CLIP Score $\uparrow$ \\ 
\midrule
  
Source Video & 0.301 & 0.583 & 0.00 & 23.21 & & 0.531 & 0.734 & 0.00 & 23.26 \\ 
\midrule

MimicBrush & 0.793 & 0.530 & \textbf{69.72} & \textbf{25.64} & & 0.890 & 0.658 & \underline{104.19} & \textbf{27.36} \\ 

AnimateDiffV3 & 0.639 & \underline{0.552} & 86.27 & 23.10 & & 0.704 & \underline{0.687} & 118.61 & 23.64 \\

AnimateDiffV3$^\star$ & \underline{0.576} & 0.538 & {71.86} & 23.74 & & \underline{0.683} & 0.665 & 111.44 & 23.43 \\

\method & \textbf{0.462} & \textbf{0.560} & \underline{71.09} & \underline{24.76} & & \textbf{0.553} & \textbf{0.710} & \textbf{96.88} & \underline{25.84} \\

\bottomrule
\end{tabular}
\renewcommand{\arraystretch}{1}
\end{adjustbox}
\caption{Quantitative comparison on the IVE-Benchmark. We assess different methods from 3 perspectives, including the temporal smoothness (Warp Error, Temporal Consistency), the visual quality of edited frames (FID) and the semantic alignment between edited frames and the reference image (CLIP Score). And we conduct experiments on two applications, texture transfer and object modification.}
\label{tab:quanticomp}
\end{table*}

\subsection{Qualitative Comparison.}
We compare our \method with baselines mentioned above.
From the results in~\cref{fig:qualicomp}, we can observe that frame-wisely editing videos through MimicBrush always results in significant discrepancy between frames since the lack of inter-frame consistency constraints; 
Additionally, due to the lack of consideration for effectively modeling semantic correlation among frames during the fine-tuning process, as well as the discrepancies between the fine-tuning process and the practical inference process, AnimateDiff and AnimateDiff$^\star$ tend to sacrifice the visual quality of the original image model editing for maintaining temporal consistency.
In contrast, our \method, by introducing \mmm for fine-tuning the model, can achieve a balance between maintaining inter-frame consistency and the visual quality of the results generated by the original image model, producing temporally smooth and visually realistic results.

\subsection{Model Analysis}

\begin{figure}[!t]
\centering
\includegraphics[width=0.99\linewidth]{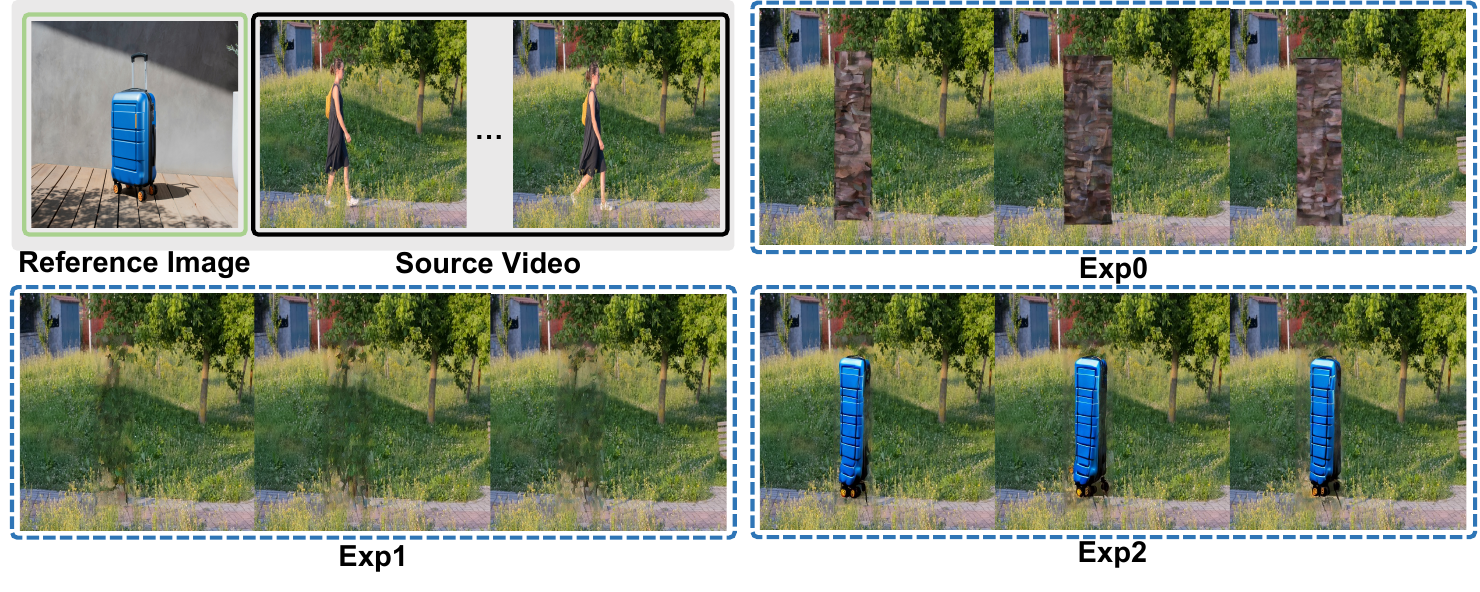}
\caption{Qualitative comparison of main components. 
}
\label{fig:main_components}
\end{figure}

\noindent\textbf{Main Components.}
We conducted a series of ablation experiments on the object modification of the IVE-Benchmark to investigate the impact of the main components in \method. And the experiments on the object modification application:
(1) \textbf{Exp0:} inflating the MimicBrush at first, and then apply the motion module AnimateDiffV3 to capture the inter-frame correlations. The fine-tuning objectives is following AnimateDiffV3.
(2) \textbf{Exp1:} introduce the \mmm to fine-tune the motion module in the model of Exp0.
(3) \textbf{Exp2:} introduce the \module into the de-noising UNet in the model of Exp1, and then simultaneously fine-tuning the \module and the motion module through \mmm. 
From \cref{tab:main_components}, we can see that the introduction of \module and \mmm can effectively enhance the base model's ability to maintain temporal consistency of edited video and preserve the visual quality of each frame. Especially, the introduction of \mmm significantly improves the performance of temporal consistency and visual quality compared to the baseline model (Exp0). 
Meanwhile, from the visualization results in \cref{fig:main_components}, we also observed some phenomena. When directly fine-tuning the motion model with the de-noising objectives instead of \mmm, the model tends to generate trivial solutions in object modification tasks (which usually involve large masked regions). 
We attribute this phenomenon to the differences between the fine-tuning and inference pipeline of the base image editing method. By using the original de-noising objective, the fine-tuning process neglects the modeling of semantic correlations between different regions of the masked input frame images, causing the model focus solely on modeling inter-frame consistency, which may harm its original ability to model semantic correlations.
Besides, when using \mmm without \module, the model sometimes over-focuses on contents outside editing regions and neglect the reference image due to inaccurate information propagation.

\begin{table}[ht]
\centering
\begin{adjustbox}{width=0.7\linewidth}
    \renewcommand{\arraystretch}{1.2}
\begin{tabular}{l|cc|cccc}
\toprule
Exps & \module & \mmm & \thead{Warp Error \\ ($10^{-1}$)}  $\downarrow$ & \thead{Temporal \\ Consistency} $\uparrow$ & FID $\downarrow$ & \thead{CLIP\\Score} $\uparrow$\\ 
\midrule
Exp0 & \textcolor{lightgray}{\ding{56}} & \textcolor{lightgray}{\ding{56}} & 0.683 & 0.665 & 111.44 & 23.43 \\
Exp1 & \textcolor{lightgray}{\ding{56}} & \ding{52} & \underline{0.558} & \underline{0.700} & \underline{99.09} & \underline{25.52} \\
Exp2 & \ding{52} & \ding{52} & \textbf{0.553} & \textbf{0.710} & \textbf{96.88} & \textbf{25.84} \\
\bottomrule
\end{tabular}
\renewcommand{\arraystretch}{1}
    \end{adjustbox}
\caption{The effect of main components in \method.}
\label{tab:main_components}
\end{table}

\noindent\textbf{Masked Motion Modeling Fine-tuning.}
We set up a series of experiments to explore the impact of mask settings in \mmm on the performance of \method, including (1) the way masks are applied to videos: using different masks for each frame or the same mask for individual video clip (\textbf{Frame-wise} and \textbf{Clip-wise} in~\cref{tab:mmm-strategy}); (2) the ratio of masked regions (\{\textbf{0, 0.25, 0.5, 0.75}\} in~\cref{tab:mmm-mr}). 
The results presented in~\cref{tab:mmm-strategy,tab:mmm-mr} indicate that by applying masks that occlude different regions to different frames within a video clip and setting a larger proportion of masked regions, the model is required to have a stronger capability for modeling both intra- and inter-frame correlations, thereby effectively enhancing the model's ability to maintain temporal consistency between frames.
For more ablation studies on the \mmm, please refer to the Appendix.

\begin{table}
\centering
\begin{adjustbox}{width=0.65\linewidth}
    \renewcommand{\arraystretch}{1.2}
\begin{tabular}{lcclcc}
\toprule
\multirow{2}{*}{Methods} &
\multicolumn{2}{c}{Texture Transfer} & & \multicolumn{2}{c}{Object Modification}
\\ \cmidrule{2-3} \cmidrule{5-6}
& \thead{Warp Error \\ ($10^{-1}$)}  $\downarrow$ & \thead{Temporal \\ Consistency} $\uparrow$ & & \thead{Warp Error \\ ($10^{-1}$)}  $\downarrow$ & \thead{Temporal \\ Consistency} $\uparrow$ \\ 
\midrule
  
0 \% & 0.509 & 0.537 & & 0.595 & 0.660 \\
25 \% & 0.469 & \underline{0.558} & & 0.558 & 0.700 \\
50 \% & \underline{0.464} & \textbf{0.560} & & \underline{0.554} & \underline{0.707} \\
75 \% & \textbf{0.462} & \textbf{0.560} & & \textbf{0.553} & \textbf{0.710} \\

\bottomrule
\end{tabular}
\renewcommand{\arraystretch}{1}
\end{adjustbox}
\caption{The effect of different masking ratio in \mmm.}
\label{tab:mmm-mr}
\end{table}

\begin{table}[h]
\centering
\begin{adjustbox}{width=0.65\linewidth}
    \renewcommand{\arraystretch}{1.2}
\begin{tabular}{l|cccc}
\toprule
Mask Strategy & \thead{Warp Error \\ ($10^{-1}$)}  $\downarrow$ & \thead{Temporal \\ Consistency} $\uparrow$ & FID $\downarrow$ & \thead{CLIP\\Score} $\uparrow$\\ 
\midrule
Clip-wise & 0.598 & 0.700 & 100.38 & 26.46 \\
Frame-wise & \textbf{0.553} & \textbf{0.710} & \textbf{96.88} & \textbf{25.84} \\
\bottomrule
\end{tabular}
\renewcommand{\arraystretch}{1}
    \end{adjustbox}
\caption{The effect of different masking strategies in \mmm.} 
\label{tab:mmm-strategy}
\end{table}

\section{Conclusion}

In this paper, we propose the first image-guided video editing diffusion model, \method. Given a source video, a reference image and masks indicating the regions to be edited in each video frame,   \method automatically edits the video by replacing masked regions with semantically related contents from the reference image.
\method is built upon a powerful image editing diffusion model. Specifically, \method firstly inflates the image editing diffusion model and then applying a motion module capable of modeling inter-frame correlations along the temporal dimension. By proposing a \module to mitigate the misleading information propagation process between masked editing frames in the source video, and a masked motion modeling fine-tuning strategy to enhance both intra- and inter-frame semantic correlations while modeling inter-frame temporal consistency, our method avoids excessive sacrifice of the visual quality of individual edited frame for maintaining inter-frame consistency.
We also constructed a benchmark (IVE-Benchmark) to assess our method from perspectives of inter-frame consistency and visual quality of edited frames.
Comprehensive experiments indicates that \method can generate temporally smooth videos while robustly handling various editing targets with high visual quality.
{
    \small
    \bibliographystyle{ieeenat_fullname}
    \bibliography{main}
}
\clearpage
\appendix
\section{IVE-Benchmark, Training and Inference Details}

\subsection{IVE-Benchmark.}
\begin{figure*}[h]
\centering
\includegraphics[width=0.99\linewidth]{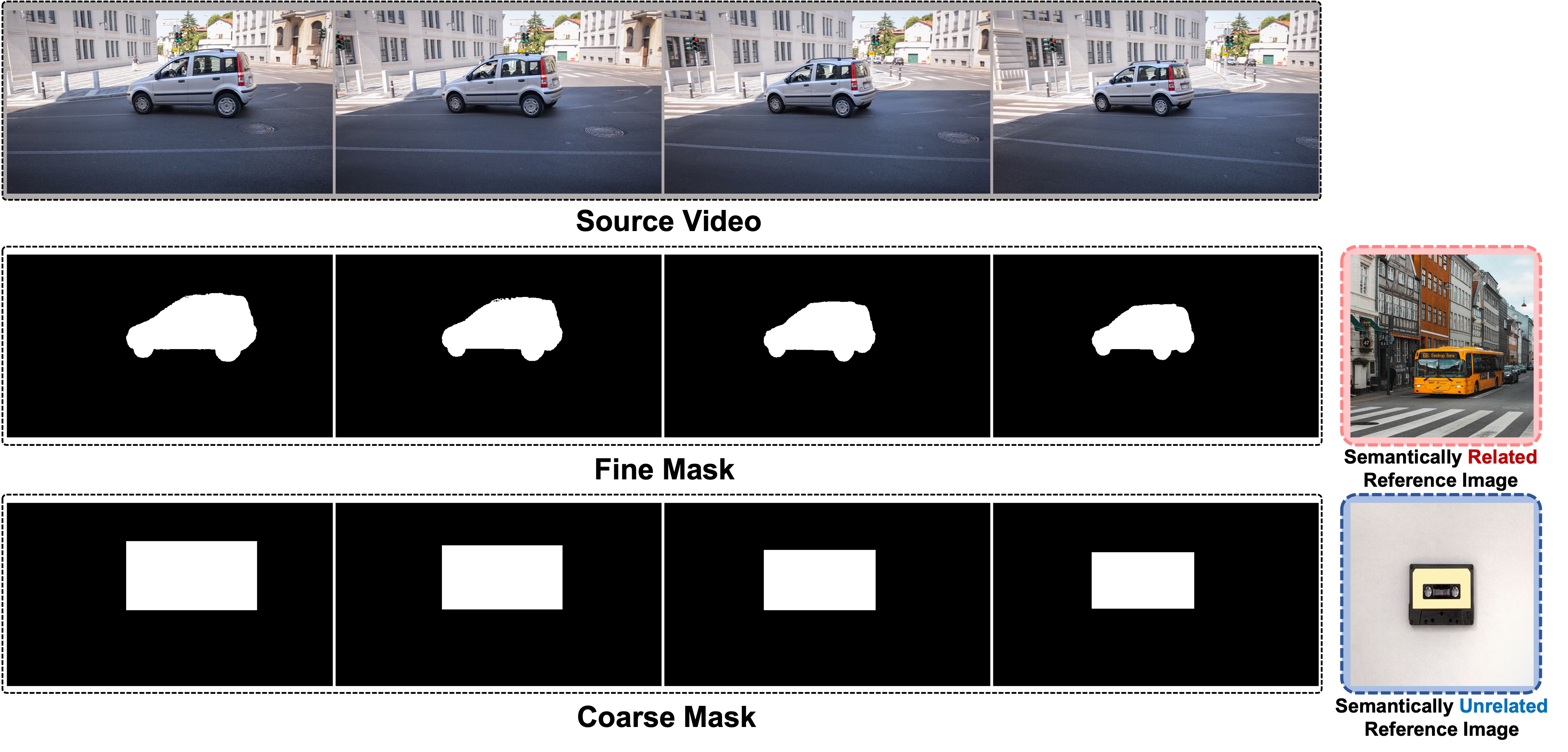}
\caption{The exemplar of \textbf{IVE-Benchmark}. For the target contents in the source video, IVE-Benchmark provides two types of mask, the fine- and coarse-grained mask for texture transfer and object modification application, respectively. Moreover, we provide two types of reference image, including the semantically related and unrelated references.
}
\label{appendix:fig-ivebench}
\end{figure*}

Similar to MimicBrush~\cite{chen2024zero}, in the construction of IVE-Benchmark, we select target editing content from videos in Davis90 and randomly found semantically similar and dissimilar reference contents from Pexels website~\cite{pexels} for texture transfer and object modification applications, respectively. For the texture transfer, we need to preserve the original shape of the object to be edited, so we used SAM2~\cite{kirillov2023segment} to obtain fine-grained masks for the specified object in each video frame; for the object modification, since the original shape of the content to be edited is not needed, we constructed rectangular masks based on fine-grained masks and set the depth map used by the model in the inference stage to guide the editing results to zero. \cref{appendix:fig-ivebench} presents examples of \textbf{\textit{video-mask-ref}} triplets. It is worth noting that for both applications, we use the same pre-trained \method and do not need to fine-tune the model for different application.

\subsection{Training Details.}
Since our chosen base image editing model, MimicBrush~\cite{chen2024zero}, was trained with the resolution of 512$\times$512, to ensure that our \method achieves the best results, we also fine-tune our model with the resolution of 512$\times$512. Limited by computational resources, we downsample the original video with a stride of 4 and select 7 frames for fine-tuning (with the first frame serving as the reference image and the rest as training data), and set the batch size to 1. All experiments are conducted with a single AMD Mi250 GPU. 
Our model is trained on the Pexels dataset following~\cite{opensoraplan}, with approximately 75,000 raw video data, for 200,000 iterations of training.

\subsection{Inference Details.}
During the inference phase, due to computational resource limitations, we edit 16 frames of images with a resolution of 336$\times$336, where 336 is the maximum resolution our GPU can handle for editing 16 frames at a time. 
\begin{table}[ht]
\centering
\begin{adjustbox}{width=0.7\linewidth}
    \renewcommand{\arraystretch}{1.2}
\begin{tabular}{l|cccc}
\toprule
Mask Strategy & \thead{Warp Error \\ ($10^{-1}$)}  $\downarrow$ & \thead{Temporal \\ Consistency} $\uparrow$ & FID $\downarrow$ & \thead{CLIP\\Score} $\uparrow$\\ 
\midrule

stride=2 & \underline{0.553} & {0.703} & 97.13 & \textbf{25.87} \\
stride=4 & \underline{0.553} & \underline{0.710} & \textbf{96.88} & \underline{25.84} \\
stride=8 & \textbf{0.549} & \textbf{0.711} & \underline{96.91} & 25.81 \\

\bottomrule
\end{tabular}
\renewcommand{\arraystretch}{1}
    \end{adjustbox}
\caption{The effect of different downsample stride in \mmm fine-tuning process.} 
\label{appendix:tab-mmm-stride}
\end{table}

\section{Ablation Study on Configurations of \mmm}

\subsection{Mask Ratio.}

We provide the complete comparison results in~\cref{appendix:tab-mmm-mr}. The results indicate that by applying masks with a larger proportion of masked regions, the model is required to have a stronger capability for modeling inter-frame dynamics and intra-frame semantic correlations, thereby effectively enhancing the model's ability to maintain temporal consistency between frames.

\begin{table*}[th]
\centering
\begin{adjustbox}{width=0.99\linewidth}
    \renewcommand{\arraystretch}{1.2}
\begin{tabular}{lcccclcccc}
\toprule
\multirow{2}{*}{Methods} &
\multicolumn{4}{c}{Texture Transfer} & & \multicolumn{4}{c}{Object Modification}
\\ \cmidrule{2-5} \cmidrule{7-10}
& Warp Error ($10^{-1}$) $\downarrow$ & Temporal Consistency $\uparrow$ & FID $\downarrow$ & CLIP Score $\uparrow$ & & Warp Error ($10^{-1}$) $\downarrow$ & Temporal Consistency $\uparrow$ & FID $\downarrow$ & CLIP Score $\uparrow$ \\ 
\midrule
  
0 \% & 0.509 & 0.537 & 73.02 & 23.14 & & 0.595 & 0.660 & 118.58 & 22.49 \\
25 \% & 0.469 & \underline{0.558} & \textbf{69.97} & 24.75 & & 0.558 & 0.700 & \underline{97.00} & 25.59 \\
50 \% & \underline{0.464} & \textbf{0.560} & \underline{70.16} & \textbf{24.84} & & \underline{0.554} & \underline{0.707} & 97.99 & \textbf{26.08}  \\
75 \% & \textbf{0.462} & \textbf{0.560} & {71.09} & \underline{24.76} & & \textbf{0.553} & \textbf{0.710} & \textbf{96.88} & \underline{25.84} \\

\bottomrule
\end{tabular}
\renewcommand{\arraystretch}{1}
\end{adjustbox}
\caption{The effect of different masking ratio in \mmm.}
\label{appendix:tab-mmm-mr}
\end{table*}

\subsection{Downsampling Stride.}
We conducted a set of experiments to analyze the downsampling stride during the training process of the original video, and the results are presented in ~\cref{appendix:tab-mmm-stride}. We observed that the model performed best when the stride was set to 4.
The reason might be that the stride influences two key factors: 
(1) The semantic correlation between the first frame serving as a reference image and the video frames used for fine-tuning. Generally, a larger stride results in a lower semantic correlation between the first frame and the subsequent training video frames. 
(2) The difficulty of modeling inter-frame correlations. 
With the stride increasing, the inter-frame correlation weakens, posing a greater challenge for \mmm in capturing these correlations. And this requires the model to possess a more robust capability to capture the motion dynamics between neighboring frames. 
Therefore, if the stride is too large, the reference image may have reduced semantic correlation with the training data, making the reconstruction task more difficult and potentially causing the model to overlook the reference information in maintaining inter-frame consistency during the fine-tuning process. On the other hand, if the stride is too short, the model might not be effectively optimized to develop its inter-frame correlation modeling capabilities, as the reconstruction process becomes too easy.

\section{Multi-Target Editing}
Since our method enables editing of fine-grained regions while completely preserving the non-edited regions unchanged, we can iteratively edit multiple targets within a single video. The results are shown in~\cref{appendix:fig-multitarget}.
\begin{figure*}[ht]
\centering
\includegraphics[width=0.99\linewidth]{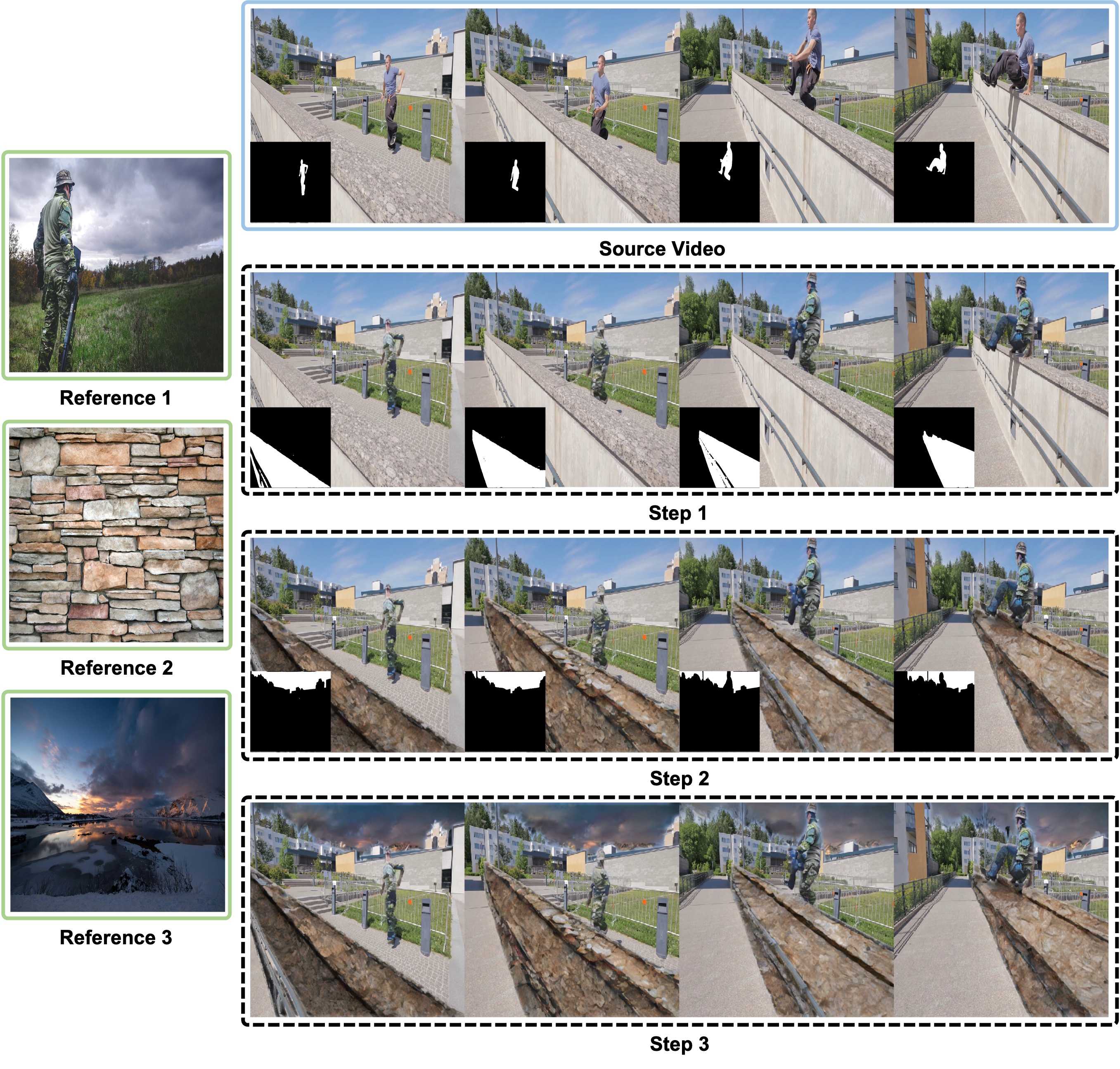}
\caption{Multi-target Editing. \textbf{[Best viewed with zoom-in.]}
}
\label{appendix:fig-multitarget}
\end{figure*}

\section{More Qualitative Comparison}
We provide more qualitative comparison between (1) frame-wisely edit the video through MimicBrush~\cite{chen2024zero}; (2) apply motion module of pretrained AnimateDiff V3~\cite{guo2023animatediff} to capture inter-frame dynamics; (3) our \method. And we provide comparisons on two editing application (texture transfer and object modification) according on the IVE-Benchmark as presented in~\cref{appendix:fig-qc_coarse,appendix:fig-qc_fine}, respectively.

\begin{figure*}[ht]
\centering
\includegraphics[width=0.99\linewidth]{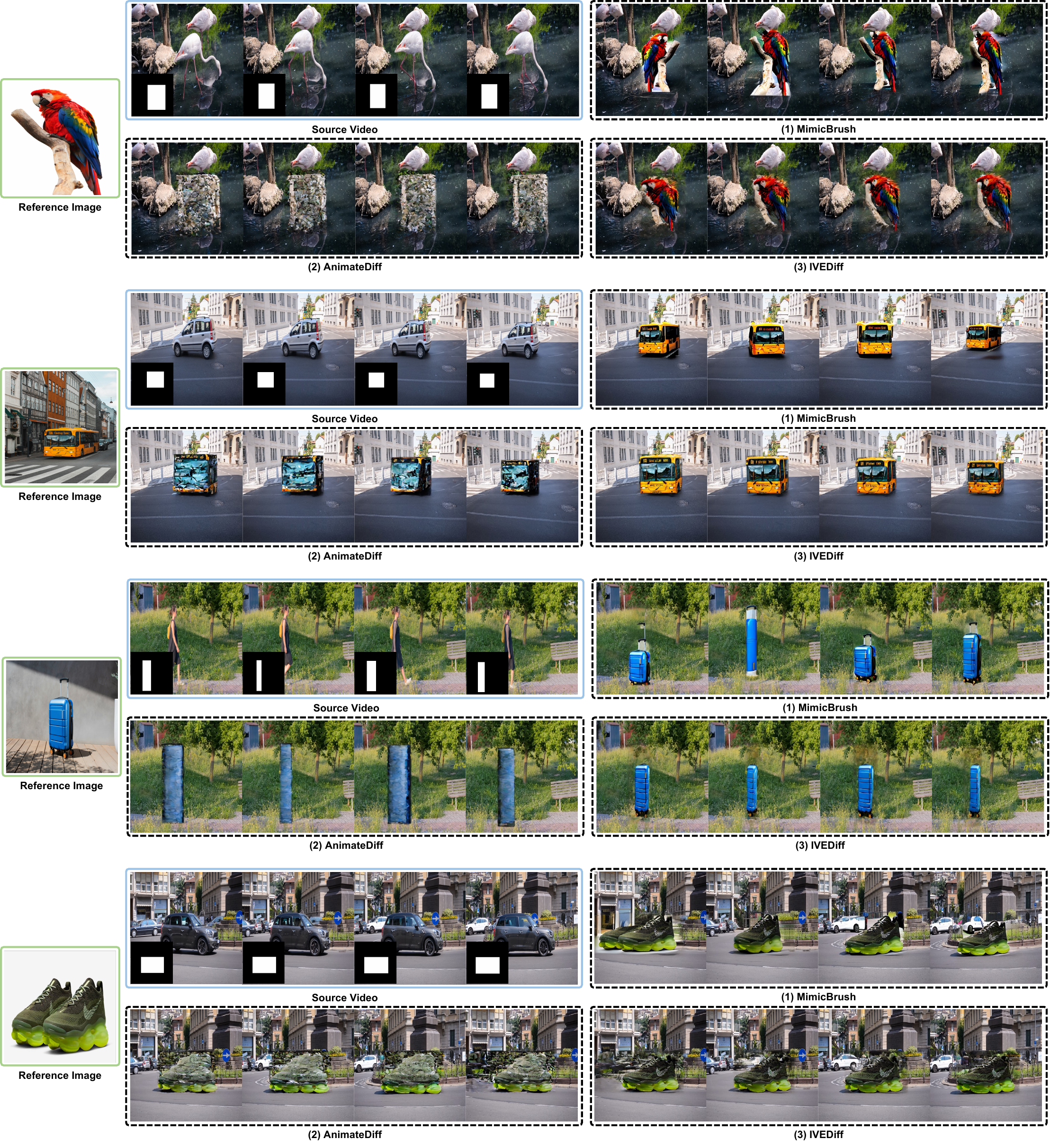}
\caption{Qualitative comparisons on the object modification application. \textbf{[Best viewed with zoom-in.]}
}
\label{appendix:fig-qc_coarse}
\end{figure*}

\begin{figure*}[ht]
\centering
\includegraphics[width=0.99\linewidth]{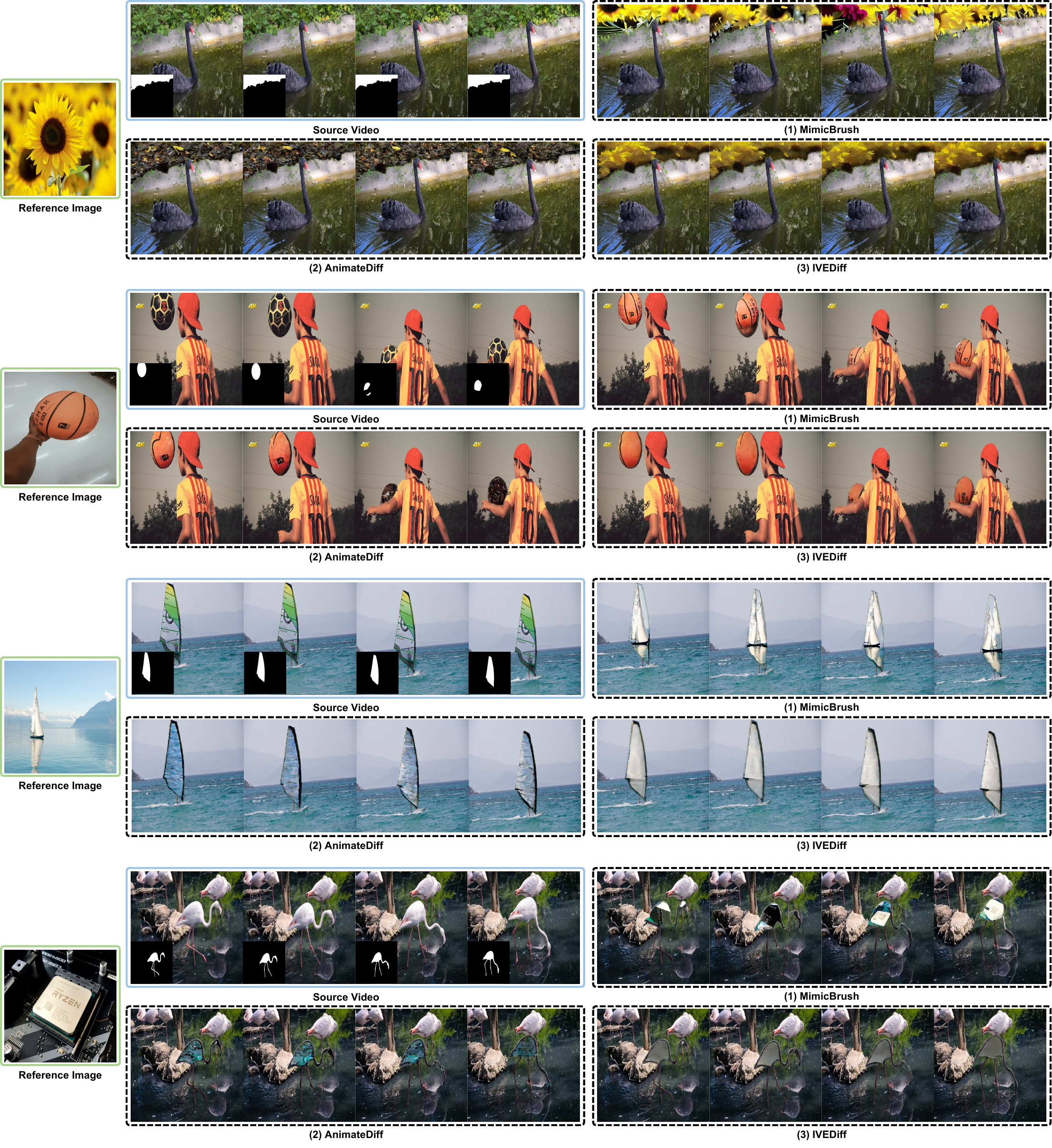}
\caption{Qualitative comparisons on the texture transfer application. \textbf{[Best viewed with zoom-in.]}
}
\label{appendix:fig-qc_fine}
\end{figure*}

\end{document}